\documentclass[pdflatex,sn-mathphys-num,iicol]{sn-jnl}

\usepackage{graphicx}%
\usepackage{multirow}%
\usepackage{amsmath,amssymb,amsfonts}%
\usepackage{amsthm}%
\usepackage{mathrsfs}%
\usepackage[title]{appendix}%
\usepackage{xcolor}%
\usepackage{soul}%
\usepackage{textcomp}%
\usepackage{manyfoot}%
\usepackage{booktabs}%
\usepackage{algorithm}%
\usepackage{algorithmicx}%
\usepackage{algpseudocode}%
\usepackage{listings}%
\usepackage{array}
\usepackage{xcolor}
\usepackage{booktabs}
\usepackage{acronym}

\acrodef{DL}{Deep Learning}
\acrodef{CNN}{Convolutional Neural Network}
\acrodef{ViT}{Vision Transformer}
\acrodef{AI}{Artificial Intelligence}
\acrodef{AGI}{Artificial General Intelligence}
\acrodef{XAI}{eXplainable Artificial Intelligence}
\acrodef{LLM}{Large Language Model}
\acrodef{PINN}{Physical Informed Neural Network}
\acrodef{ML}{Machine Learning}
\acrodef{CAM}{Class Activation Map}

\theoremstyle{thmstyleone}%
\theoremstyle{thmstyletwo}%

\theoremstyle{thmstylethree}%

\begin{document}

\title[Article Title]{Batch-CAM: Introduction to better reasoning in convolutional deep learning models}


\author*[1,2]{\fnm{Giacomo} \sur{Ignesti}}\email{giacomo.ignesti@isti.cnr.it}

\author[1]{\fnm{Davide} \sur{Moroni}}\email{davide.moroni@isti.cnr.it}
\equalcont{These authors contributed equally to this work.}
\author[1]{\fnm{Massimo} \sur{Martinelli}}\email{massimo.martinelli@isti.cnr.it}
\equalcont{These authors contributed equally to this work.}

\affil*[1]{\orgdiv{ISTI}, \orgname{CNR}, \orgaddress{\street{Via Giuseppe Moruzzi}, \city{Pisa}, \postcode{56124},\country{Italy}}}

\affil*[2]{\orgdiv{Computer Science Department}, \orgname{University of Pisa}, \orgaddress{\street{Largo Bruno Pontecorvo, 3}, \city{Pisa}, \postcode{56127}, \country{Italy}}}

\abstract{Deep learning opacity often impedes deployment in high-stakes domains. We propose a training framework that aligns model focus with class-representative features without requiring pixel-level annotations. 
To this end, we introduce Batch-CAM, a vectorised implementation of Gradient-weighted Class Activation Mapping that integrates directly into the training loop with minimal computational overhead. We propose two regularisation terms: a Prototype Loss, which aligns individual-sample attention with the global class average, and a Batch-CAM Loss, which enforces consistency within a training batch. These are evaluated using $L_1$, $L_2$, and SSIM metrics. 
Validated on MNIST and Fashion-MNIST using ResNet18 and ConvNeXt-V2, our method generates significantly more coherent and human-interpretable saliency maps compared to baselines. While maintaining competitive classification accuracy, the framework successfully suppresses spurious feature activation, as evidenced by qualitative reconstruction analysis. 
Batch-CAM appears to offer a scalable pathway for training intrinsically interpretable models by leveraging batch-level statistics to guide feature extraction, effectively bridging the gap between predictive performance and explainability.
}


\maketitle

\section{Introduction}\label{sec1}
Although modern \ac{DL} models for computer vision, notably \acp{CNN} \cite{woo2023ConvNeXt} \acp{ViT} \cite{dosovitskiy2020image}, have demonstrated exceptional performance, their black-box nature fundamentally limits our understanding of their decision-making processes. This opacity, a direct consequence of their intricate, non-linear optimisation, poses significant challenges for model interpretability, accountability, and safe deployment in high-stakes scenarios such as medical diagnostics, financial modelling, or autonomous driving. For instance, a model trained to detect pneumonia might achieve high accuracy by focusing on a hospital-specific pattern, while being statistically effective on the training data. However, such a model would be erroneous when deployed in a new hospital \cite{ye2024spurious}. The divergence between a model's high predictive accuracy and its lack of human-interpretable reasoning creates a gap that must be addressed for the broader, responsible adoption of \ac{AI}. \ac{AGI} exemplifies this challenge, aiming to develop models whose generalisation capabilities extend beyond their training domain, driven by an ability to simulate genuine thinking \cite{AGI2023}. To address this, \ac{XAI} provides techniques to make \ac{AI} systems transparent and comprehensible, fostering trust and exposing vulnerabilities. The extensive development of \ac{XAI} techniques \cite{Bodria2023} is just one possible approach to understanding \ac{AI} and developing more robust models. Other approaches focus on designing novel model architectures and training paradigms that inherently enhance the reasoning and generalisation capabilities of \ac{DL}. A modern example of this evolution is the emergence of \acp{LLM}, which are predominantly built upon the Transformer architecture \cite{zhao2023survey}. Unlike their predecessors, which were often designed for specific, narrow tasks, \acp{LLM} represent a shift towards more generalised intelligence. 

\begin{figure*}[h] 
\centering
\includegraphics[width=0.8\textwidth]{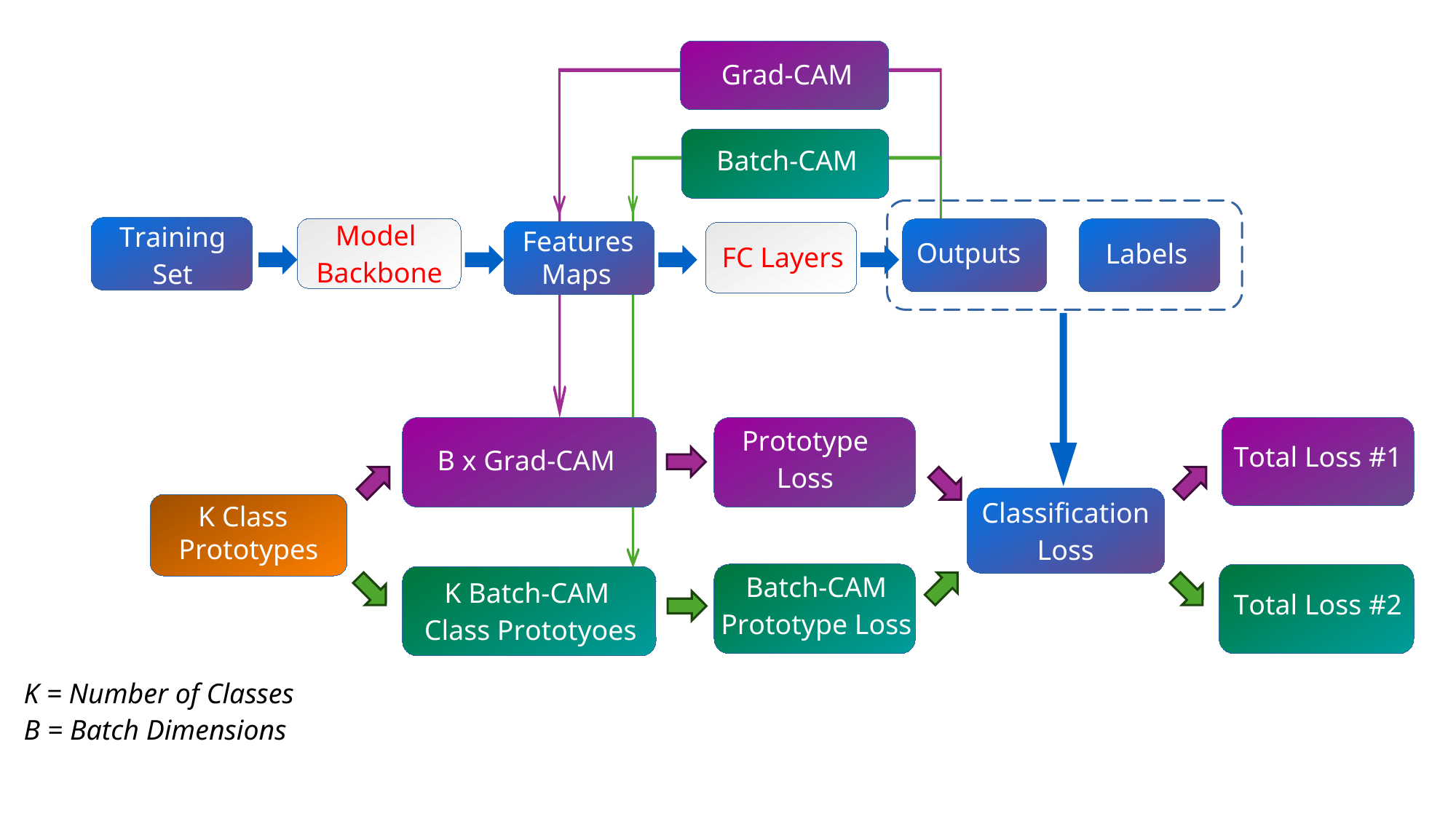}
\caption{Overview of the proposed Batch-CAM architecture and the integrated training pipeline. The diagram illustrates the flow from the input training set through the model backbone and feature extraction stages. Localisation is performed via standard Grad-CAM and the proposed Batch-CAM modules. The learning objective is defined by a composite total loss function according to two approaches. Namely,  the classification loss, derived from the model outputs and labels, is combined either with the specialized prototype losses $\mathcal{L}_{\text{Proto}}^{\phi}$  and  $\mathcal{L}_{\text{Batch}}^{\phi}$ generated from class-specific prototypes as described in the following of the paper (Section \ref{sec33}).} \label{mod:arch}
\end{figure*}
This paper expands a novel training paradigm \cite{ignesti2024,ignesti2024you} to enhance computer vision by integrating an explanatory mechanism into the learning process itself. Our core hypothesis is that the model already possesses sufficient information within the training data to be both precise and capable of generalisation, provided it is guided to focus on the correct features. We introduce a method that combines the Grad-CAM algorithm \cite{selvaraju2017grad} with a tailored reconstruction loss term. This forces the model not only to classify an image correctly but also to detect which part of the image should get its focus. By comparing this generated explanation against a class prototype, our loss function provides evidence on which part of the image should be used to return the classification, effectively compelling it to learn for the right reasons. We extend this concept by exploring two innovative approaches: an image-to-prototype comparison to learn from a generalised class distribution and a Batch-CAM technique that regularises the model's averaged focus across a batch.
We introduce two new primary innovations to this end:
\begin{enumerate}
    \item Image-to-Prototype Comparison: A mechanism that aligns the focus of individual samples with a generalised class distribution.
    \item Batch-CAM Regularisation: A technique designed to stabilise and regularise model focus at the batch level 
\end{enumerate}
The resulting pipeline (as shown in Figure \ref{mod:arch}) embeds an eXplainable AI (XAI)-driven feedback loop within a standard Convolutional Neural Network (CNN) architecture. The framework is evaluated using robust distance metrics, including $L_1$ and $L_2$ losses, as well as the Structural Similarity Index (SSIM) loss. Our results demonstrate that this spatial supervision significantly improves both model generalisation and interpretability.
Section \ref{sec2} explores the foundational concepts and related works that underpin our contribution. Section \ref{sec3} details the evolution of our proposed architecture, explaining the algorithmic improvements and the different approaches investigated. Section \ref{sec4} presents a comprehensive performance evaluation on several benchmark datasets, including MNIST and Fashion-MNIST, followed by an in-depth analysis of the results. Finally, Section \ref{sec5} concludes by proposing future research directions that aim to extend and validate our approach in more complex domains. This paper extends and refines the previous workshop paper \cite{ignesti2024,ignesti2024you}.

\section{Foundations and Related Works}\label{sec2}
The challenge of creating trustworthy AI has inspired the development of a range of interpretability techniques. Our work is situated at the intersection of three key research topics: post-hoc explanation methods that analyse trained models, intrinsically interpretable models designed for transparency, and training-time interventions that guide a model's learning process. These steps are achieved by manipulating the loss function, leveraging XAI algorithms, and utilising the training process efficiently.

\subsection{The Loss Function as a Control Mechanism}
While standard loss functions like Cross-Entropy remain staples for \ac{DL} optimisation \cite{foret2020sharpness}, the increasing complexity of \ac{AI} necessitates more sophisticated objective functions \cite{Terven2025}. Advanced models often employ composite losses to enforce diverse constraints. For instance, DINOv3 integrates cross-entropy with specific regularisers for feature uniformity and non-degradation \cite{simeoni2025dinov3}, demonstrating that loss functions can be engineered to shape internal representations beyond mere accuracy. Another approach to loss control is presented in \acp{PINN}\cite{raissi2017physics}, in which the input data are correlated with the data through a bound in the loss term. A restraint can be a boundary condition for a hypothetical physical phenomenon whose violation would be impossible under physical laws. \acp{PINN} have evolved beyond simple forward problems to tackle complex inverse tasks, such as inferring physical parameters from limited data. As reviewed by Karniadakis et al. \cite{karniadakis2021physics}, their application in domains ranging from biomedical engineering to digital twins demonstrates their efficacy in imbuing models with robust domain knowledge.

\subsection{Paradigms in Explainable AI}
\ac{XAI} methods can be broadly categorised by when and how they provide explanations. The two primary families are post-hoc techniques, which explain decisions after a model is trained, and intrinsically interpretable models, which are transparent by design. Post hoc methods are the most common approach to interpreting existing black-box models. They operate by analysing a trained model to understand its behaviour for a specific input. One of the earliest and most influential methods is LIME (Local Interpretable Model-agnostic Explanations), which approximates the complex decision boundary of any model in the local vicinity of a prediction with a simpler, linear model \cite{ribeiro2016should}.
In computer vision, attribution methods are dominant. These techniques generate saliency maps (or heatmaps) that highlight the pixels or regions of an input image most influential to the final prediction. \ac{CAM} and its widely used successors, Grad-CAM and Grad-CAM++ \cite{chattopadhay2018grad} are pioneering examples. Grad-CAM employs final-layer gradients to localize critical image regions. However, while essential for diagnosing spurious correlations, such methods are purely diagnostic and lack a direct mechanism for correction.
In contrast to post-hoc analysis, other research lines seek to build models that are transparent by design. Instead of explaining a decision after the fact, the model's reasoning process is inherently understandable to humans. In computer vision, prototype-based networks are a prominent example \cite{chen2019looks}. These models learn a set of representative exemplar archetypes for each class during training. To make a prediction, the model compares parts of the input image to its learned prototypes and explains its decision by highlighting the similarities between these parts and the prototypes. While offering powerful reasoning, this approach may compromise predictive performance and prove inflexible for tasks ill-suited to part-based comparisons \cite{hoffmann2021looks}.

\subsection{Guiding Models to Learn for the ``Right Reasons''}
A third paradigm synthesises these ideas by employing interpretability tools not merely for analysis, but as an active component of the training process itself. The objective is to constrain the model, ensuring it bases decisions on valid, human-approved evidence. The work by Ross et al. represents a foundational effort in this direction \cite{rightreason}. They proposed penalising the model's input gradients during training; by adding a term to the loss function that suppresses sensitivity to features pre-identified by humans as irrelevant, they could guide the learning process. The primary limitation, however, is the significant overhead of requiring additional human annotations to explicitly mark these irrelevant features. Subsequent research has sought to refine this approach. For instance, later studies proposed using influence functions rather than input gradients to trace predictions back to training data, offering a more robust measure of feature importance \cite{shao2021right}. Similarly, methods such as Contextual Decomposition Explanation Penalisation were introduced to incorporate prior knowledge as an explanation error in the loss function, penalising the model if its internal attributions deviate from the desired logic \cite{rieger2020interpretations}. While powerful, these methods generally rely on the availability of prior knowledge or additional annotations to guide the model.

Our work aligns with this paradigm but seeks to mitigate the reliance on explicit human supervision regarding irrelevant features. By leveraging class prototypes as weak targets for the model's attention (as visualised via Grad-CAM), we aim to provide a supervisory signal derived directly from the data distribution itself. This establishes a more scalable and automated method for training models that learn for the right reasons.

\section{Methods}\label{sec3}
The proposed method is evaluated on the MNIST and F-MNIST datasets. This algorithm represents an evolution of the \textit{Contrastive Loss Approach} previously proposed in \cite{ignesti2024,ignesti2024you}. 

In those earlier studies, the training process utilised the Grad-CAM algorithm to guide image classification towards a significant feature space, ensuring alignment between the model's focus and the relevant parts of the input images. However, the gradient-accessing mechanism employed relied on software hooks, which introduced computational inefficiencies (as discussed in Section \ref{sec33}). Furthermore, the regularisation strategy was inherently contrastive: the loss function minimised the bi-dimensional Euclidean distance between the generated Grad-CAM maps and the input images. This formulation compared the activation maps directly to the specific inputs, rather than the more generalised class prototypes introduced in the current work.

The total loss in the previous framework combined a contrastive (C) reconstruction term with standard classification loss, defined as follows:

\begin{equation}
\label{eq:1}
L_{\text{C}} =\frac{1}{2}\left( Y \cdot D^2 + (1-Y) \cdot \max(m - D, 0)^2 \right)
\end{equation}

\noindent where:
$Y$ is the label ($1$ for similar pairs, $0$ for dissimilar pairs);
$D$ is the Euclidean distance between the two output tensors;
$m$ is the margin (a hyperparameter).

The Multi-Class Cross-Entropy Loss ($L_{\mathrm{CE}}$) was expressed as:

\begin{equation}
\label{eq:2}
L_{\mathrm{CE}} = -\frac{1}{N} \sum_{i=1}^{N} \left( Y_i \log(p_i) + (1 - Y_i) \log(1 - p_i) \right)
\end{equation}

The loss function proposed in the following section retains the goal of guided attention but diverges from this precedent by removing the contrastive component and the reliance on inefficient hooks.

\subsection{Data}\label{sec31}
The data are initially analysed using the one-channel approach, and the only data transformation is tensorisation and normalisation to the one-channel MNIST and F-MNIST literature values for mean and standard deviation (equal to  [0.1307, 0.3081] and [0.2860, 0.3530], respectively). No data augmentation techniques were employed, except resizing the input images as detailed in Section \ref{sec:exp}.
The prototypes class images are constructed using the training set and validation set of both datasets. The images are obtained by averaging over these two sets, and the reconstructed prototypes are used separately during the training and validation steps of the model. This method uses the gradient during the validation step, but does not update it, as it should be done only in the training step to avoid contamination errors.

\subsection{Model}\label{sec32}
Three model architectures are tested to benchmark the proposed new algorithms. The first is a baseline CNN that processes an image and flattens its features into a vector before a fully connected layer classifies it into one of the 10 classes of MNIST or Fashion-MNIST.
The other two, ResNet18 and ConvNeXt-V2-Tiny, were chosen to assess the algorithms' effectiveness on more advanced architectures known for their robust feature extraction capabilities.
ResNet \cite{he2016deep} introduces residual connections, while ConvNeXt-V2 introduces techniques as a fully convolutional masked autoencoder (FCMAE) framework and a new Global Response Normalisation (GRN) \cite{woo2023ConvNeXt}.

\subsection{Reconstruction GradCAM}\label{sec33}

In previous works \cite{ignesti2024,ignesti2024you}, reconstruction was performed sequentially during the training process on an image-by-image basis. Identifying this as a computational bottleneck, the entire reconstruction process is redesigned in this paper. Instead of analysing instances individually, Grad-CAM is evaluated at the end of each batch using a vectorisation approach, which significantly accelerates the process. The primary distinction lies in the mechanism and timing of gradient computation and extraction.

Previously, in the hook-based approach, CAM generation was a \textit{stateful} process that relied on architectural side-effects; it required registering hooks to intercept forward and backward passes, storing intermediate tensors in global variables ($\mathcal{A}$ and $\mathcal{G}$), and utilising computationally expensive operations like \texttt{retain\_graph=True}. This introduced significant memory overhead and instability by decoupling the explanation from the model's primary execution. In contrast, the new architecture redefines CAM generation as a \textit{stateless} operation, denoted as $M = \Phi(L, A)$. By treating the explanation as a localised partial derivative, Equation \ref{eq:pdev}, computed via direct gradient calculation, such as \texttt{torch.autograd.grad}, where $S_c$ is the score for the target class. This formulation ensures that the XAI loop is fully side-effect-free on the primary backward pass. In this way, it also avoids the overhead of full-model backpropagation for each image.

\begin{equation}
\label{eq:pdev}
    G = \frac{\partial S_c}{\partial A}
\end{equation}
Explicitly passing Logits ($L$) and Feature Maps ($A$) as inputs, the new method integrates the explanation as a primary variable within the unified computational graph. This eliminates the dependency on external hooks, streamlines gradient flow, and ensures optimal resource utilisation within the training loop. By computing only the precise gradients needed for the feature maps $A$, the method achieves significantly higher throughput, enabling the use of complex explanation-based regularisation even on large-scale datasets. 
By redefining Grad-CAM as a functional component of the loss pipeline, we transition the role of XAI from a "passive observer" to an "active supervisor" within the optimisation process. This shift necessitates a computationally efficient formulation that can handle multidimensional data streams. In the following sections, we detail the implementation of the proposed algorithm and contrast it with the traditional approach. Specifically, we analyse the processing of a batch size $N$ and feature maps $A \in \mathbb{R}^{N \times K \times H \times W}$, where $K$, $H$, and $W$ denote the number of channels, height, and width, respectively.
\subsection{Proposed Loss}\label{sec331}
The Loss approach has also been updated so that explanations for all images of the same class should be semantically similar. The model isn't asked to reconstruct the original input image from its CAM. Instead, it's trained to create an explanation for any given class instance that resembles an abstract prototype for that class. The prototype is calculated once at the beginning by averaging all training and validation images belonging to a specific class (e.g., the prototype for a T-shirt, a Number, or a medical condition). The goal is to enforce consistency in what the model is classifying. The two proposed losses are:
\begin{enumerate}
\item{Prototype Loss (Per-Image Consistency): This loss operates on an individual level. For each image in a batch, it generates a Grad-CAM and calculates the loss between that CAM and the pre-computed prototype for the image's actual class.}

\item{Batch-CAM Prototype Loss (Batch-Level Consistency): This is a more aggregated approach. Instead of generating a CAM for every single image, it computes the average CAM per class present in the batch. It does this by averaging the gradients and feature maps for all instances of that class. This single, batch-averaged CAM is then compared to the class prototype using the loss. This encourages the model to generate consistent explanations for a class at a group level. 
To compute the Batch-CAM for a given class $c$, the feature maps from all $N_c$ samples belonging to that class within batch $B$ are averaged (Equation \ref{eq:avg_feature_map}). Similarly, the gradients of the class score with respect to the feature maps are averaged across all $N_c$ samples (Equation \ref{eq:avg_gradients}). Finally, the neuron importance weights are calculated by applying Global Average Pooling (GAP) to the averaged gradients (Equation \ref{eq:neuron_importance}).
\begin{align}
\label{eq:avg_feature_map}
\bar{A}_k^c &= \frac{1}{N_c} \sum_{i \in \mathcal{B}_c} A_{i,k} \\
\label{eq:avg_gradients}
\bar{G}_k^c &= \frac{1}{N_c} \sum_{i \in \mathcal{B}_c} \frac{\partial S_c}{\partial A_{i,k}} \\
\label{eq:neuron_importance}
\bar{\alpha}_k^c &= \frac{1}{H \times W} \sum_{u=1}^{H} \sum_{v=1}^{W} (\bar{G}_k^c)_{u,v}
\end{align}}

\end{enumerate}

The loss function can be formulated using various metrics. Although prior work employed Euclidean distance ($L_2$ norm) to quantify reconstruction error, we identified this approach as a performance bottleneck. Consequently, this study presents a comparative analysis of classical $L_1$ and $L_2$ norm-based reconstruction losses, alongside SSIM, to determine the most effective objective for our training paradigm.

The Structural Similarity Index (SSIM) Loss relies on the degradation of structural information between two images. It operates on a single scale, synthesizing three core components: luminance ($\mathcal{L}$), contrast ($\mathcal{C}$), and structure ($\mathcal{S}$). The total loss is computed as one minus the SSIM index (Equation \ref{eq:ssim_loss}).

\begin{equation}
\label{eq:ssim_loss}
\mathcal{L}_{\text{SSIM}}(x, y) = 1 - [\mathcal{L}(x, y)]^{\alpha} \cdot [\mathcal{C}(x, y)]^{\beta} \cdot [\mathcal{S}(x, y)]^{\gamma}
\end{equation}

Where:
\begin{itemize}
    \item $x$ and $y$ are the two images being compared.
    \item $\mathcal{L}, \mathcal{C}, \mathcal{S}$ are the luminance, contrast, and structure comparisons, respectively (Equations \ref{eq:l}-\ref{eq:s}).
    \begin{align}
        \mathcal{L}(x, y) &= \frac{2\mu_x\mu_y + \epsilon_1}{\mu_x^2 + \mu_y^2 + \epsilon_1} \label{eq:l} \\
        \mathcal{C}(x, y) &= \frac{2\sigma_x\sigma_y + \epsilon_2}{\sigma_x^2 + \sigma_y^2 + \epsilon_2} \label{eq:c} \\
        \mathcal{S}(x, y) &= \frac{\sigma_{xy} + \epsilon_3}{\sigma_x\sigma_y + \epsilon_3} \label{eq:s}
    \end{align}
    Here, \(\mu_{x,y}\) and \(\sigma_{x,y}\) represent the pixel mean and standard deviation of the images. The terms \(\epsilon_{1,2,3}\) are small positive constants included to avoid division by zero (instability).
    \item $\alpha, \beta, \gamma$ are weighting parameters used to adjust the relative importance of each component.
\end{itemize}

In our framework, the total objective function combines the standard Cross-Entropy Loss ($\mathcal{L}_{\text{Class}}$) with a regularization term derived from either the Prototype Loss or the Batch-CAM Prototype Loss.

When utilizing the \textbf{Prototype Loss} (Per-Image Consistency), we define an alignment term $\mathcal{L}_{\text{Proto}}^{\phi}(M_i, P_c)$, where $M_i$ represents the Grad-CAM for image $i$, $P_c$ denotes the prototype for its corresponding class $c$, and $\phi$ represents the chosen metric ($L_1$, $L_2$, or SSIM).

The total loss for this configuration is formulated as:
\begin{equation}
\label{eq:total_loss_per_image_proto}
\mathcal{L}_{\text{Total}} = \lambda_{\text{class}} \mathcal{L}_{\text{Class}} + \lambda_{\text{proto}} \mathcal{L}_{\text{Proto}}^{\phi}(M,P)
\end{equation}

Where:
\begin{itemize}
    \item $\mathcal{L}_{\text{Class}}$ is the standard Cross-Entropy Loss for classification.
    \item $\mathcal{L}_{\text{Proto}}^{\phi}$ represents the alignment loss calculated using the chosen metric ($L_1$, $L_2$, or SSIM) between the per-image Grad-CAMs ($M$) and their respective class prototypes ($P$).
    \item $\lambda_{\text{class}}$ and $\lambda_{\text{proto}}$ are scalar hyperparameters that balance the contribution of each loss component.
\end{itemize}

Alternatively, for the \textbf{Batch-CAM Prototype Loss} (Batch-Level Consistency), we enforce alignment between the \textit{class-averaged} attention maps and their corresponding prototypes. We denote this regularizer as $\mathcal{L}_{\text{Batch}}^{\phi}(\bar{M}_c, P_c)$, where $\bar{M}_c$ is the averaged Grad-CAM for all samples of class $c$ within the batch.

The total objective function is defined as:
\begin{equation}
\label{eq:total_loss_batch}
\mathcal{L}_{\text{Total}} = \lambda_{\text{class}} \mathcal{L}_{\text{Class}} + \lambda_{\text{batch}} \sum_{c \in C_{\mathcal{B}}} \mathcal{L}_{\text{Batch}}^{\phi}(\bar{M}_c, P_c)
\end{equation}

Where:
\begin{itemize}
    \item $C_{\mathcal{B}}$ represents the set of unique classes present in the current training batch.
    \item $\mathcal{L}_{\text{Batch}}^{\phi}$ quantifies the alignment between the batch-averaged Grad-CAM ($\bar{M}_c$) and the class prototype ($P_c$) using the chosen metric $\phi$.
    \item $\lambda_{\text{batch}}$ is the hyperparameter controlling the strength of this batch-level regularization.
\end{itemize}

\subsection{Experiments}\label{sec:exp}
Our experimental evaluation proceeded in two stages. First, we benchmarked three architectures (Simple CNN, ResNet-18, ConvNeXt-V2-Tiny) on $28\times28$ inputs across seven training conditions: a standard Cross-Entropy baseline and six variations of our proposed Prototype and Batch-CAM losses. Observing that the resulting $3\times3$ feature maps in deep networks were insufficient for detailed visual analysis, we advanced to a second stage. Here, we retrained the best-performing ResNet-18 and ConvNeXt-V2-Tiny models on $112\times112$ inputs, allowing for high-resolution Grad-CAM generation and a deeper analysis of the model's learned spatial reasoning.

\section{Results}\label{sec4}

\subsection{Quantitative Performance}
We first evaluate the classification performance of the proposed training paradigms. As summarized in Table \ref{tab:best_accuracy_exp}, models trained with Batch-CAM Prototype Loss (BCPL) achieve accuracy comparable to, and in some cases marginally higher than, the baseline models trained with standard Cross-Entropy. For instance, ConvNeXt-V2 on MNIST reached 99.72\% accuracy with BCPL ($L_1$), matching the baseline's 99.71\%. Similarly, ResNet18 on Fashion-MNIST showed a slight improvement (94.99\% vs. 94.82\%). 
A computational cost analysis indicates that our functional implementation of Grad-CAM introduces negligible overhead. The number of parameters remains constant, and the training GFLOPs are virtually identical to the baseline, validating the efficiency of the proposed vectorised implementation.
\begin{table}[h!]
\centering
\caption{Best accuracy and corresponding experiment for each model on $28\times28$ inputs. BCPL: Batch-CAM Prototype Loss.}
\label{tab:best_accuracy_exp}
\footnotesize 
\setlength{\tabcolsep}{1.5pt}
\begin{tabular}{@{}lcccc@{}} 
\toprule
\multirow{2}{*}{\textbf{Model}} & \multicolumn{2}{c}{\textbf{MNIST}} & \multicolumn{2}{c}{\textbf{F-MNIST}} \\ 
\cmidrule(lr){2-3} \cmidrule(l){4-5} 
& \textbf{Exp.} & \textbf{Acc.(\%)} & \textbf{Exp.} & \textbf{Acc.(\%)} \\
\midrule
\shortstack[l]{ConvNeXt-\\V2 Tiny} & BCPL\_$L_1$ & 99.72 & Baseline & 94.79 \\
\addlinespace 
ResNet18 & BCPL\_$L_2$ & 99.65 & BCPL\_$L_2$ & 94.99 \\
SimpleCNN & BCPL\_$L_1$ & 99.00 & BCPL\_$L_1$ & 91.32 \\
\bottomrule
\end{tabular}
\end{table}

\subsection{Qualitative Analysis and Reconstruction}
\begin{figure}[h]
\centering
\includegraphics[width=\columnwidth]{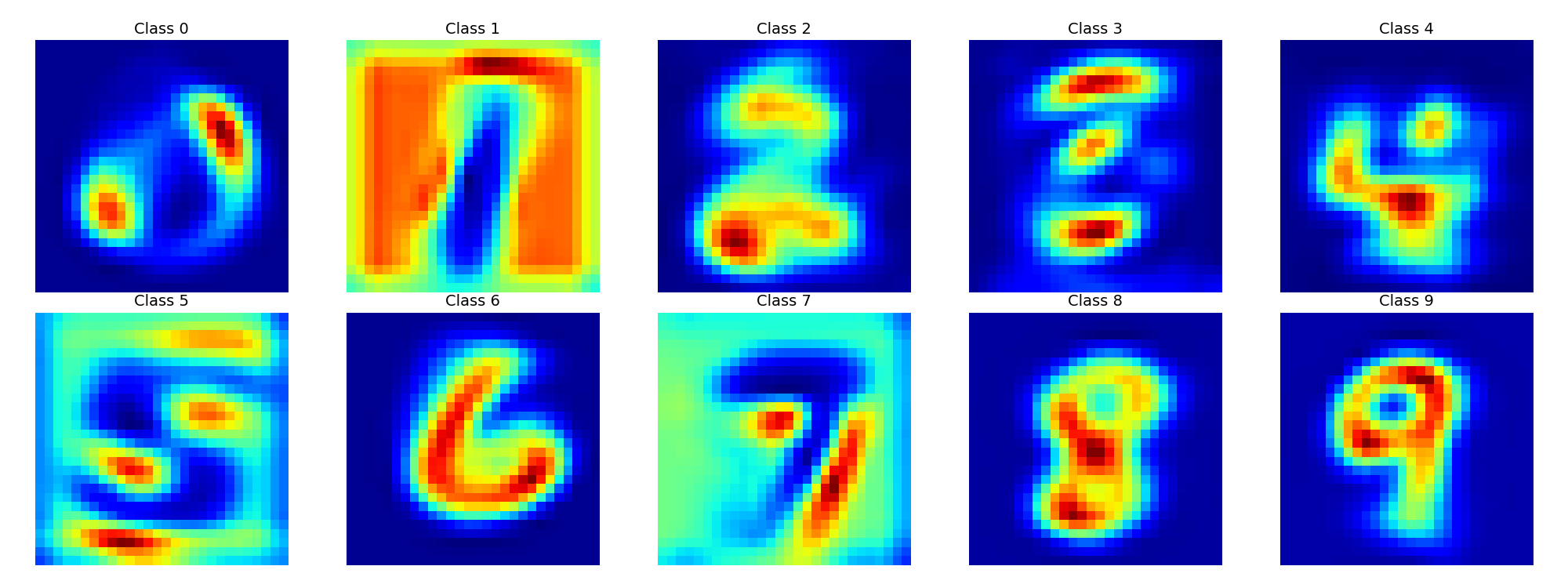}
\caption{Average MNIST reconstruction prototypes per class ($28\times28$ inputs) for a CNN trained with standard Cross-Entropy loss. Note the lack of structural definition.} \label{mod:size}
\end{figure}

While classification metrics remain stable, significant differences emerge in the model's internal reasoning, as evidenced by the generated prototypes.
Standard CNNs typically learn robust classifiers but fail to maintain spatial coherence in their feature maps. As shown in Figure \ref{mod:size}, the activation maps produced by the baseline model are low-resolution proxies that poorly reconstruct the original input.

\begin{figure}[h]
\centering
\includegraphics[width=\columnwidth]{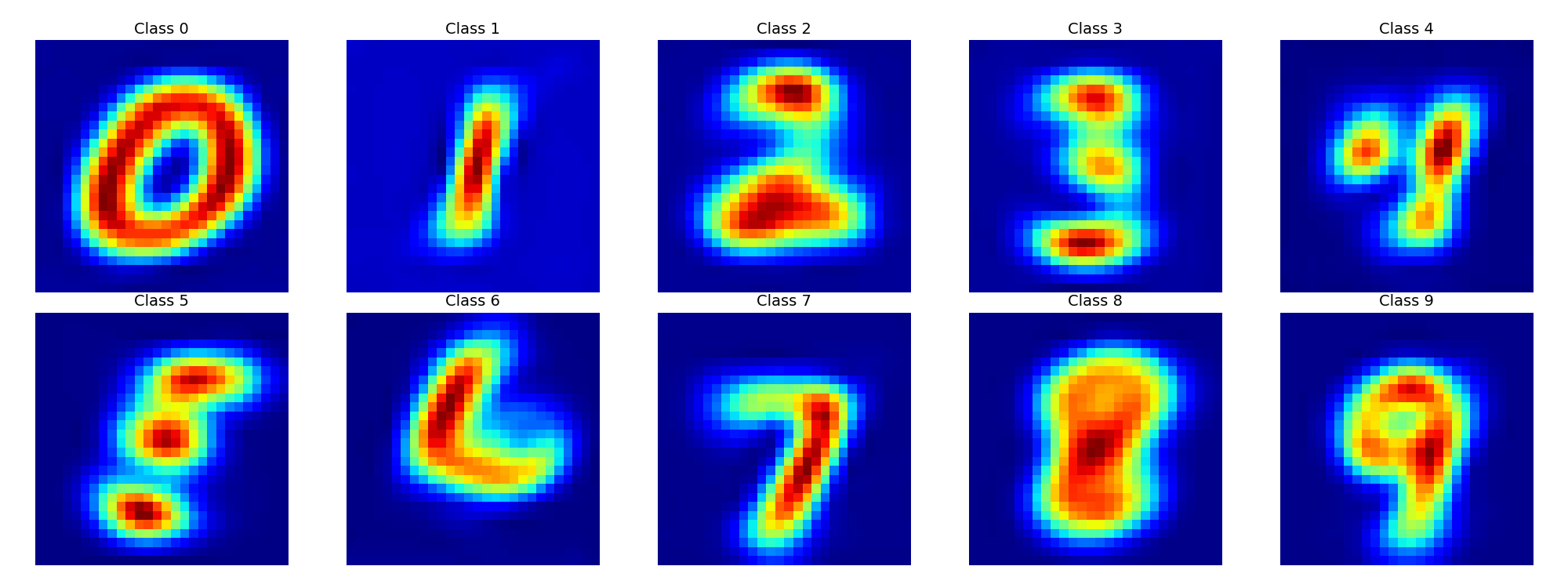}
\caption{Average MNIST reconstruction prototypes per class ($28\times28$ inputs) for a CNN trained with Batch-CAM Prototype Loss. The prototypes exhibit clear structural alignment with the digits.} \label{mod:right}
\end{figure}

In contrast, models trained with BCPL demonstrate a stronger alignment between the activation maps and the class archetypes. Figure \ref{mod:right} illustrates that the Batch-CAM guided models produce prototypes that are not only semantically consistent but also generalize well to the test set.

\subsubsection{High-Resolution Analysis}
\begin{figure}[h]
\centering
\includegraphics[width=\columnwidth]{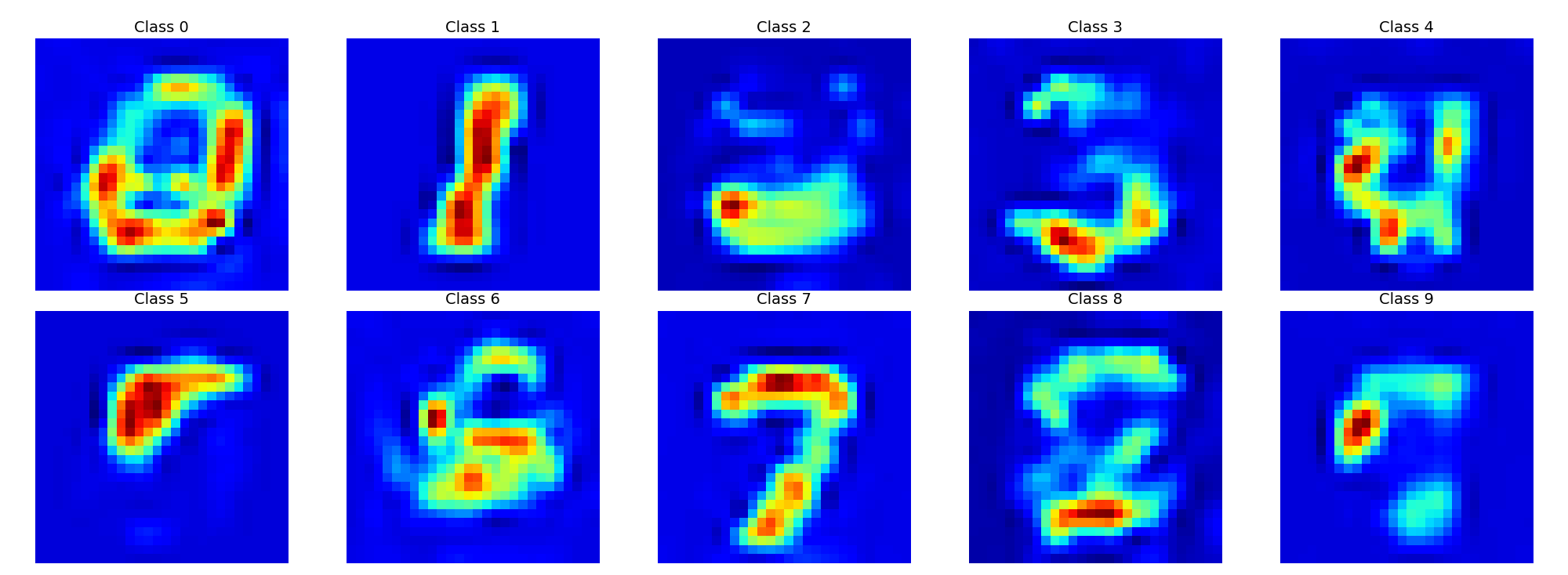}
\caption{Reconstructed prototypes for MNIST ($112\times112$) using ConvNeXt-V2 with BCPL ($L_1$).} \label{mod:mnistc}
\end{figure}

\begin{figure}[h]
\centering
\includegraphics[width=\columnwidth]{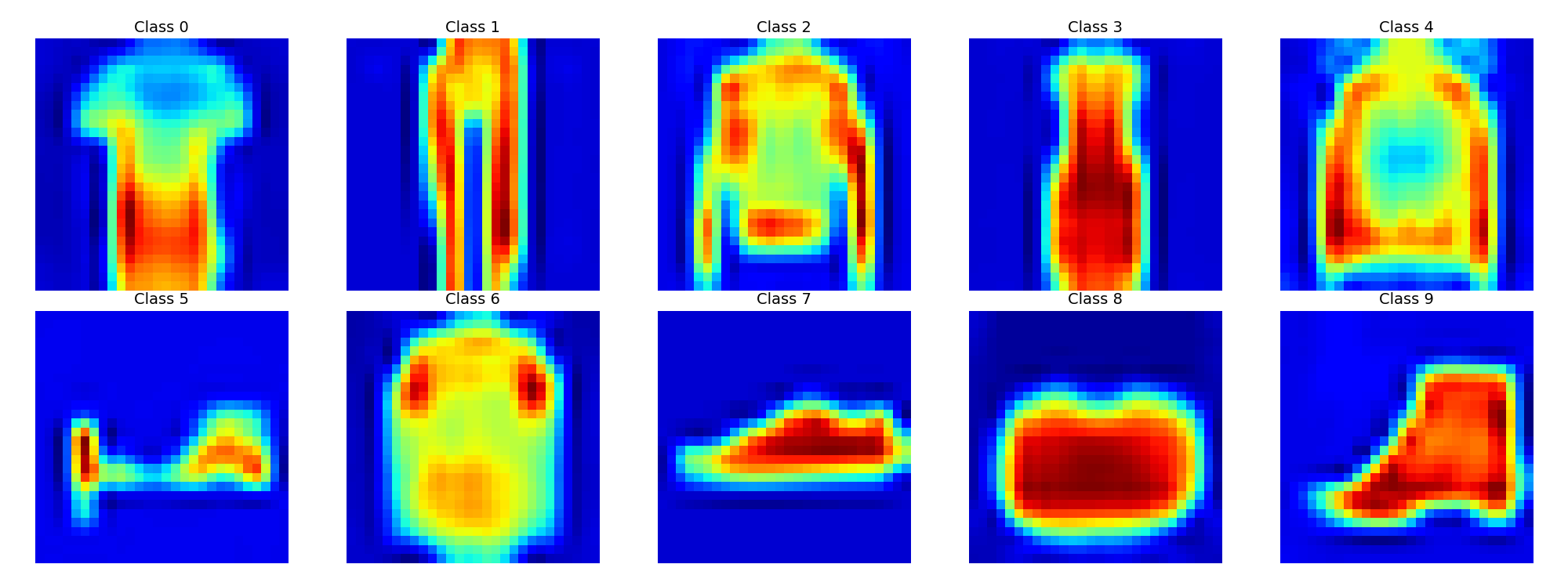}
\caption{Reconstructed prototypes for Fashion-MNIST ($112\times112$) using ConvNeXt-V2 with BCPL ($L_1$).} \label{mod:fmnistc}
\end{figure}

The retraining of ResNet18 and ConvNeXt-V2 on $112\times112$ images (Table \ref{tab:best_accuracy_112}) allows for a more detailed inspection of feature attention. Figures \ref{mod:mnistc} and \ref{mod:fmnistc} confirm that the models focus on defining characteristics of the objects (e.g., the loop of a digit or the silhouette of a garment) rather than background noise.

\begin{table}[h!]
\centering
\caption{Best model performance retrained on $112\times112$ inputs.}
\label{tab:best_accuracy_112}
\footnotesize
\setlength{\tabcolsep}{1.5pt}
\begin{tabular}{@{}lcccc@{}}
\toprule
\multirow{2}{*}{\textbf{Model}} & \multicolumn{2}{c}{\textbf{MNIST}} & \multicolumn{2}{c}{\textbf{F-MNIST}} \\
\cmidrule(lr){2-3} \cmidrule(l){4-5}
& \textbf{Exp.} & \textbf{Acc.(\%)} & \textbf{Exp.} & \textbf{Acc.(\%)} \\
\midrule
\shortstack[l]{ConvNeXt-\\V2 Tiny} & BCPL\_$L_1$ & 99.74 & BCPL\_$L_1$ & 95.01\\
\addlinespace
ResNet18 & BCPL\_$L_1$ & 99.71 & BCPL\_$L_1$ & 94.99 \\
\bottomrule
\end{tabular}
\end{table}

\subsubsection{Error Analysis via Prototypes}
\begin{figure}[h]
\centering
\includegraphics[width=\columnwidth]{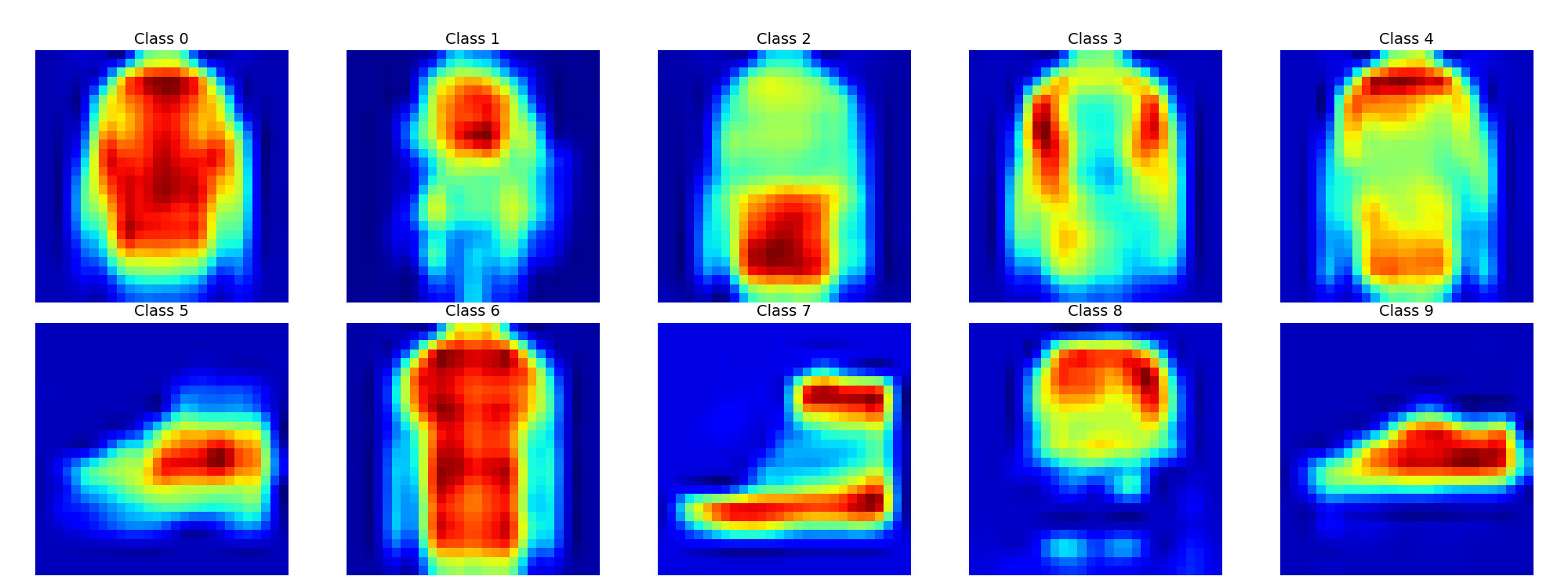}
\caption{Average prototypes for misclassified images (Fashion-MNIST, $112\times112$). The distorted attention maps reveal the specific features that caused the confusion.} \label{mod:wrngF-MNIST}
\end{figure}

The reconstructed prototypes also serve as a diagnostic tool for misclassifications. Figure \ref{mod:wrngF-MNIST} visualises the attention maps for erroneous predictions. For example, when the model misclassifies a Pullover (Class 2), the activation map shifts from the torso and sleeves to a non-descript region at the bottom, resembling the profile of a Coat (Class 4). Similarly, confusion between Ankle boots (Class 9) and Sneakers (Class 7) is characterized by a focus on the sole rather than the distinguishing ankle structure. This transparency allows for the identification of specific failure modes in the model's reasoning process.

\section{Conclusion}\label{sec5}
This paper presented Batch-CAM, a training paradigm that integrates a vectorised Grad-CAM implementation directly into the learning objective. By regularising the model to align its spatial attention with class-level prototypes, we demonstrated that it is possible to enhance interpretability without compromising classification accuracy. The architectural shift from hook-based to functional gradient computation ensures that this explainability comes with minimal computational cost.

The approach currently exhibits some limitations. The method relies on averaging feature maps to generate class prototypes, which implicitly assumes a unimodal distribution (i.e., that all instances of a class share a similar spatial structure). While effective for aligned datasets like MNIST and Fashion-MNIST, this assumption may not hold for complex, in-the-wild datasets where intra-class variance is high (e.g., objects appearing in different poses or scales). In such cases, a simple average may yield a blurry or (even) meaningless prototype. Future work will address this by exploring, for instance, clustering-based prototype generation to handle multi-modal distributions and extending the validation to more complex datasets. In any case, the work provides a proof of concept that explainability can be effectively treated as a supervised learning objective.


\section*{Statements and Declarations}

\begin{itemize}
    \item Funding: The authors did not receive support from any organization for the submitted work. No funding was received for conducting this study.
    \item Competing Interests: The authors have no relevant interests, financial and non, to disclose.
    \item Author Contributions: All authors contributed to the study conception and design.
    \item Data Availability: The datasets analyzed during the current study (MNIST and Fashion-MNIST) are available in open-access repositories. No other data were used in this study.
    \item Ethics Approval: Not applicable.
    \item Consent to Participate: Not applicable.
\end{itemize}

\bibliography{sn-bibliography}

\end{document}